\documentclass{article}
\usepackage[final]{corl_2026}
\usepackage[T1]{fontenc}
\usepackage{amsmath,amssymb,graphicx,booktabs,array,needspace}
\usepackage{xurl} % allow long reference URLs to break anywhere
\newcommand{\pihalf}{\ensuremath{\pi_{0.5}}}
\makeatletter
\renewcommand{\@conferencelocation}{Austin, Texas, USA}
\renewcommand{\@notice}{\@float{noticebox}[b]\footnotesize\@noticestring\end@float}
\makeatother
\title{Can Astra Solve RoboMME\\without Breaking the Bank?}
\author{%
  Bingao Chen \quad Haoquan Fang \quad C. Karen Liu\\
  Stanford University\\
  \normalfont\small\href{https://bingaochen.github.io/Astra-on-RoboMME/}{\nolinkurl{bingaochen.github.io/Astra-on-RoboMME/}}%
}
\hypersetup{pdftitle={Can Astra Solve RoboMME without Breaking the Bank?},pdfauthor={Bingao Chen, Haoquan Fang, C. Karen Liu},pdfsubject={CoRL 2026 Workshop on Memory for Robot Foundation Models}}
\begin{document}
\maketitle
\begin{abstract}
Robotic manipulation often requires remembering information that is no longer visible, such as an object's location before occlusion or the order of a demonstrated action sequence. A large vision-language planner can read this history, but calling it before every action chunk is slow and costly. We present a three-tier system for the RoboMME benchmark. GPT-6 Astra (System~2) selects a grounded subtask from stored visual history, a \pihalf{} vision-language-action policy (System~1) executes it, and a fine-tuned Qwen3-VL-4B monitor (System~1.5) checks after each action chunk whether the subtask is finished. The instruction is kept until the monitor reports completion or a task-specific event occurs, while the policy keeps acting on new observations. On RoboMME, the system succeeds in 79.13\% of episodes with an average of 3.63 Astra calls and 80.84 seconds per episode. This is 34.62 points above the best published non-oracle baseline and 4.95 points below GroundSG+Oracle, which gives the low-level policy ground-truth subgoals.
\end{abstract}

\section{Introduction}
Manipulation can depend on information that is no longer visible: an object's identity before it is covered, the order of a demonstrated sequence, or the number of repetitions already completed. RoboMME~\citep{dai2026robomme} tests these dependencies with 16 tasks in four suites: Counting, Permanence, Reference, and Imitation. A common design for such tasks is hierarchical, with a vision-language model (VLM) choosing the next subtask and a vision-language-action (VLA) policy executing it.

In our development runs, small high-level VLMs often misread RoboMME's video inputs and issued wrong subtasks, so we use GPT-6 Astra~\citep{openai2026astra} as the planner. Calling it before every action chunk, however, is expensive and mostly redundant: a chosen subtask usually remains valid for several chunks, and whether it has finished can be judged from the last few frames.

We therefore place a small learned monitor between the planner and the policy. Astra is called at the start of an episode and again only when the monitor reports completion or a task-specific event occurs. In between, the VLA keeps executing the same instruction from fresh observations. We evaluate the full system on RoboMME and report success rates, planner usage, and failure cases.

\paragraph{Related work.}
RoboMME compares symbolic and perceptual memory interfaces~\citep{dai2026robomme}. SAM2Act adds a visual memory bank~\citep{fang2025sam2act}, MemER retrieves keyframes~\citep{sridhar2026memer}, and MEM combines short-horizon video with long-horizon text memory~\citep{torne2026mem}. Hierarchical systems such as $\tau_0$-VLA use execution memory and test-time reasoning to select subtasks~\citep{cai2026tau}. Our system instead sends task-dependent history to an external planner at each call and lets a local monitor decide when to call it.

\section{Three-Tier System}
\label{sec:method}
\begin{figure}[t]
\centering
\includegraphics[width=\linewidth]{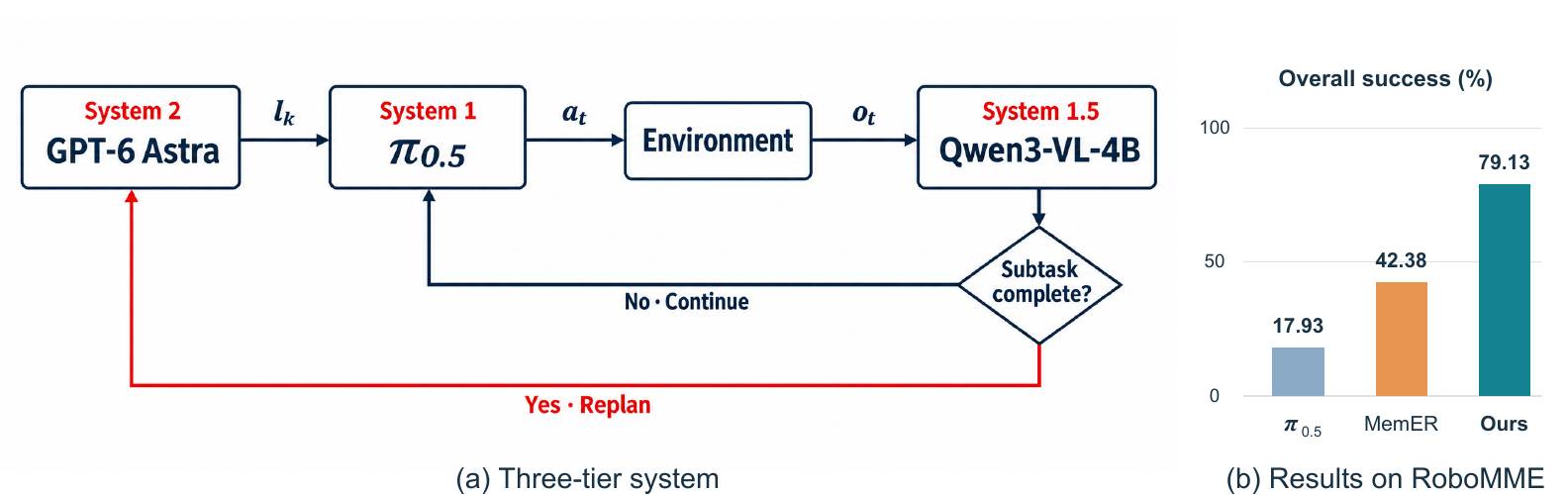}
\caption{\textbf{Planning, action, and monitoring at different frequencies.} (a) Astra uses task history to choose a subtask instruction $l_k$. The VLA executes actions $a_t$ in 16-step chunks; the environment returns observations $o_t$, and a local monitor checks progress after each chunk. The instruction is kept until an event requires a new high-level decision (Section~\ref{sec:coordination}). (b) Overall success on RoboMME compared with two published baselines from Table~\ref{tab:results}.}
\label{fig:method}
\end{figure}

Figure~\ref{fig:method}(a) shows the three components. Let $g$ be the episode instruction, $o_t$ the observation at environment step $t$, and $l_k$ the $k$th grounded subtask.

\subsection{System 2: History and Grounded Planning}
At each planning event, Astra produces one executable instruction:
\begin{equation}
 l_k\sim\pi_{\mathrm{HL}}(\,\cdot\mid o_t,m_t,g\,),
\end{equation}
where $m_t$ is the history passed to the planner. It has three parts: the demonstration video when the task provides one, selected execution frames that may show objects now hidden, and a text progress record that lists issued subtasks and monitor-confirmed subtasks separately. Repeated actions are recorded as separate entries, so their order and count remain available. Each request also includes the current front-camera image and the image at execution start. Astra keeps no state between calls; all history is sent with each request.

Which history is sent and which output templates are allowed are configured per task: nine tasks receive the demonstration video and four receive dense execution history (Appendix~\ref{app:implementation}; full prompts in Appendix~\ref{app:prompts}). When a target must be localized, the instruction contains a pixel coordinate in the current image. The output is checked against the templates before it reaches the VLA.

\subsection{System 1: Low-Level Action Execution}
We use RoboMME's grounded-subgoal \pihalf{} policy~\citep{black2025pi05}:
\begin{equation}
 a_{t:t+H-1}\sim\pi_{\mathrm{LL}}(\,\cdot\mid o_t,l_k,g\,),\qquad H=16,
\end{equation}
where $o_t$ includes camera images and robot state. Each chunk runs 16 environment steps. The same $l_k$ can guide several consecutive chunks, and each chunk is predicted from a new observation, so control remains closed-loop while Astra is idle.

\subsection{System 1.5: Visual Progress Monitoring}
\label{sec:monitor}
The monitor is Qwen3-VL-4B-Instruct~\citep{qwen3vl}, fine-tuned with LoRA on RoboMME demonstrations to predict a binary event for the active subtask:
\begin{equation}
 p_{\mathrm{MON}}(e_t\mid v_t,l_k,g), \qquad e_t\in\{0,1\}.
\end{equation}
The input $v_t$ has ten images: eight recent front-camera frames taken three steps apart (spanning up to 21 steps), the front image from when the current instruction was issued, and the current wrist image. This reference image is reset whenever an instruction is issued, even if its text repeats an earlier one, so each repetition is compared with its own starting state. In most tasks the event marks completion. In StopCube it signals an approaching target, and in ButtonUnmaskSwap Astra verifies a reported completion of the second button press (Appendix~\ref{app:events}).

\subsection{Event-Triggered Coordination}
\label{sec:coordination}
\begin{figure}[t]
\centering
\includegraphics[width=\linewidth]{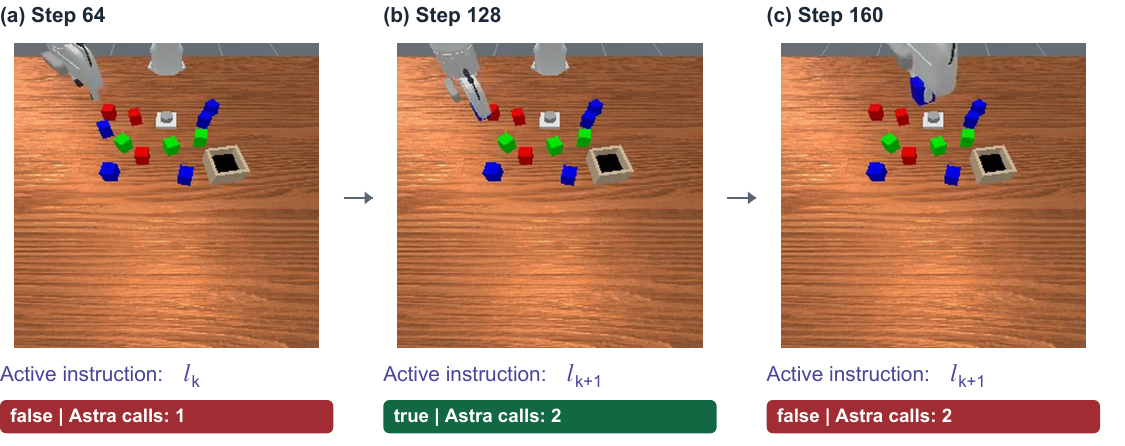}
\caption{\textbf{Instruction reuse between planning events.} Three frames from a successful recorded episode. A negative check at step 64 keeps $l_k$. Completion at step 128 triggers Astra's second call and the next instruction $l_{k+1}$; the positive check shown refers to $l_k$. At step 160, a negative check continues $l_{k+1}$. Details are in Appendix~\ref{app:visuals}.}
\label{fig:execution-compact}
\end{figure}

Astra is called once at initialization, and the monitor runs after every action chunk. A negative decision keeps the current instruction for another chunk. A completion event adds the subtask to the progress record and requests the next one. The StopCube timing event and the ButtonUnmaskSwap verification event send the execution history to Astra before the system advances. Figure~\ref{fig:execution-compact} shows instruction reuse across one completion event; Appendix~\ref{app:visuals} gives full walkthroughs.

\paragraph{Inference scheduling.}
As in RoboMME's official two-system evaluation, the simulator pauses while Astra runs, so planning time adds to wall-clock duration but not to the step budget. In preliminary runs without pausing, the VLA used up the 1,300-step budget in most episodes before Astra returned.

\section{Experiments}
\label{sec:experiments}
\paragraph{Setup.}
We evaluate on the official RoboMME test set (50 episodes per task) with a 1,300-step budget per episode and report simulator task success. Baseline numbers are the published RoboMME results~\citep{dai2026robomme}: \pihalf{}~\citep{black2025pi05} with the current observation only; FrameSamp+Modul and SAM2Act+~\citep{fang2025sam2act} with perceptual memory; and MemER~\citep{sridhar2026memer}, GroundSG+QwenVL, and GroundSG+Gemini with a high-level planner. The last two use a fine-tuned Qwen3-VL-4B~\citep{qwen3vl} and a prompted Gemini~2.5 Pro~\citep{gemini25}. GroundSG+Oracle gives ground-truth subgoals to the low-level policy and serves as a reference for grounded execution. Astra uses medium reasoning effort; the VLA, monitor, and simulator share one NVIDIA L40S GPU.

\begin{table}[t]
\centering
\caption{\textbf{Success rates (\%) on RoboMME.} Baselines are published RoboMME results~\citep{dai2026robomme}. GroundSG+Oracle uses ground-truth subgoals. Bold marks the best non-oracle result.}
\label{tab:results}
\small
\setlength{\tabcolsep}{4pt}
\begin{tabular}{lrrrrr}
\toprule
Method & Counting & Permanence & Reference & Imitation & Overall \\
\midrule
\pihalf{}~\citep{black2025pi05} & 28.78 & 17.00 & 17.16 & 8.78 & 17.93 \\
FrameSamp+Modul~\citep{dai2026robomme} & 65.22 & 25.11 & 36.33 & 51.39 & 44.51 \\
SAM2Act+~\citep{fang2025sam2act} & 35.33 & 26.00 & 16.83 & 7.33 & 21.37 \\
MemER~\citep{sridhar2026memer} & 48.83 & 53.17 & 38.00 & 29.50 & 42.38 \\
GroundSG+QwenVL~\citep{dai2026robomme} & 38.00 & 39.34 & 31.56 & 21.89 & 32.70 \\
GroundSG+Gemini~\citep{dai2026robomme} & 12.75 & 15.75 & 11.25 & 6.50 & 11.56 \\
\textbf{Three-tier Astra (ours)} & \textbf{69.50} & \textbf{94.00} & \textbf{92.00} & \textbf{61.00} & \textbf{79.13} \\
\midrule
GroundSG+Oracle~\citep{dai2026robomme} & 83.86 & 93.31 & 95.17 & 63.98 & 84.08 \\
\bottomrule
\end{tabular}
\end{table}

\paragraph{Task success.}
The system reaches 79.13\% overall success (Table~\ref{tab:results}): 94.00\% on Permanence, 92.00\% on Reference, 69.50\% on Counting, and 61.00\% on Imitation. Overall success is 34.62 points above FrameSamp+Modul, the best non-oracle baseline, and 4.95 points below GroundSG+Oracle. The gap to the oracle is largest in Counting (14.36 points, versus 3.17 in Reference and 2.98 in Imitation), and it comes mostly from StopCube: the other three Counting tasks average 86.67\%.

\paragraph{Failure cases.}
Of the 167 unsuccessful episodes, 103 fail and 64 time out. StopCube and InsertPeg each succeed in only 9/50 episodes and together account for 82 of them. StopCube requires pressing a button on a specific passage of a moving cube; all 41 of its unsuccessful episodes are failures rather than timeouts. InsertPeg requires contact-rich insertion; 36 of its 41 unsuccessful episodes time out. We have not yet attributed these failures to planning, monitoring, or execution errors. Appendix~\ref{app:results} lists outcomes for all 16 tasks.

\paragraph{Planner usage.}
The system averages 3.63 Astra calls per episode, counting both planning and visual-verification requests, and 80.84 seconds of wall-clock time. The recorded trajectories contain 24.62 action chunks on average, so calling Astra before every chunk would require about 24.62 calls along the same trajectories. Writing these averages as $\bar C_{\mathrm{Astra}}$ and $\bar C_{\mathrm{chunk}}$,
\begin{equation}
 1-\frac{\bar C_{\mathrm{Astra}}}{\bar C_{\mathrm{chunk}}}
 =1-\frac{3.63}{24.62}\approx85.3\%.
\end{equation}
This figure assumes the recorded trajectories; a controller that replans at every chunk would follow different trajectories and could reach a different success rate. Call count is also only a proxy for cost, because request size grows with the supplied history and the monitor runs after every chunk.

\section{Limitations and Conclusion}
\paragraph{Limitations.}
We evaluate only the full system. Ablations that remove the history, replace Astra with a smaller planner, or replace the monitor with fixed-interval replanning are needed to separate the contributions of memory, planner capability, and monitoring. The GroundSG baselines differ from our system in planner, inputs, and replanning schedule, so they are not matched comparisons. We did not run an always-replan controller or measure token usage and monetary cost. We also did not measure the monitor's closed-loop error rate; a false completion corrupts the progress record used by later calls. History selection, output templates, and event rules are set per task, which limits transfer to new tasks. Finally, the simulator pauses during planning, so we have not tested scenes that change while Astra reasons.

\paragraph{Conclusion.}
On RoboMME, the three-tier system reached 79.13\% success while calling Astra 3.63 times per episode, compared with 24.62 action chunks. The ablations and cost measurements above would show which parts of the system account for this result.

\clearpage
\bibliography{references}
\clearpage
\appendix
\raggedbottom
\section*{Appendix}
\section{Visual Walkthroughs}
\label{app:visuals}
The following figures show selected frames from two successful hard test episodes. Each execution panel pairs the front-camera view with the active Astra instruction, the latest monitor decision, and the cumulative number of Astra calls.

\begin{figure}[h]
\centering
\includegraphics[width=\linewidth]{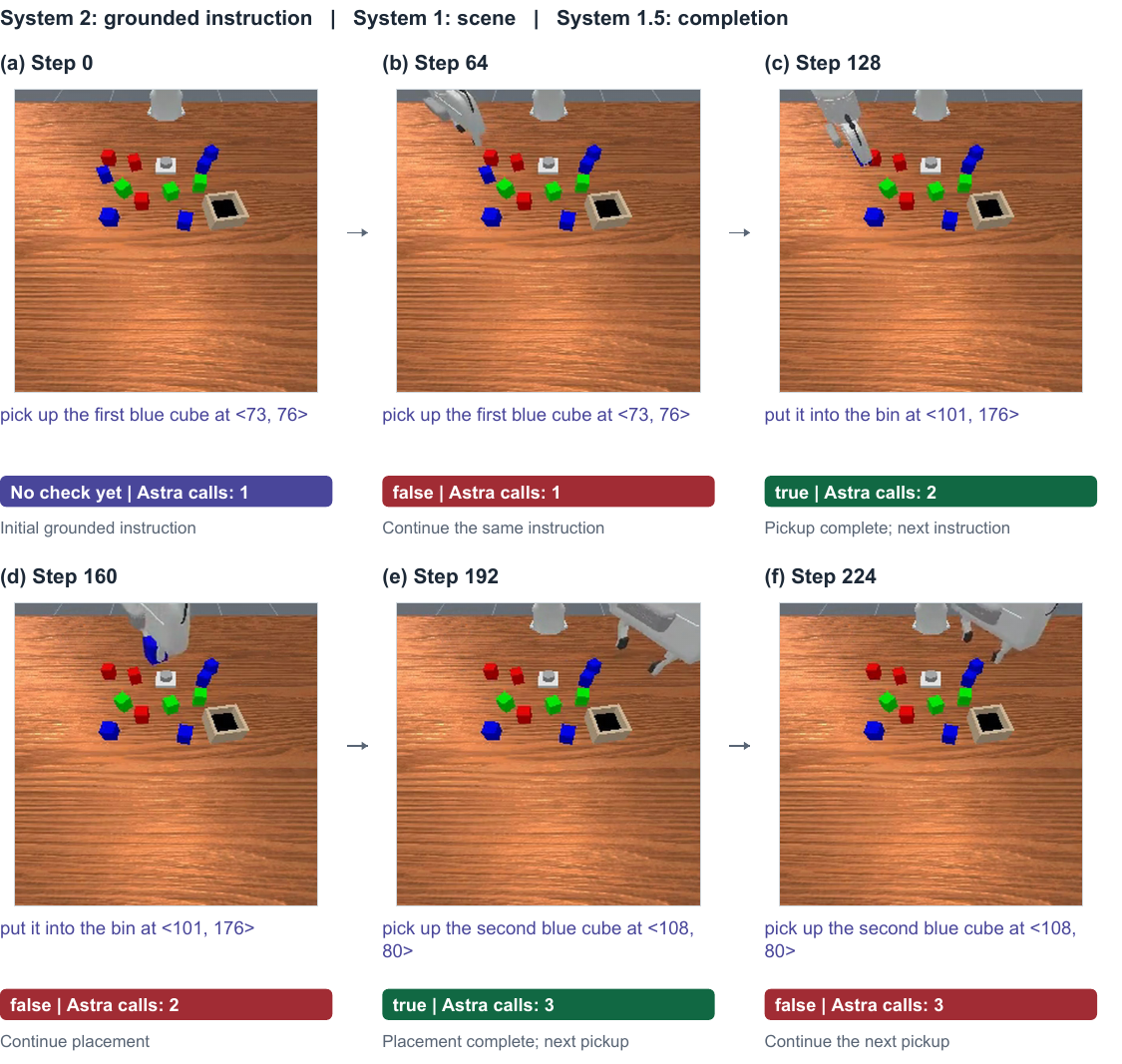}
\caption{\textbf{Planning, action, and monitoring during execution.} BinFill, hard test episode 023. The first pickup instruction is kept at steps 0 and 64. Completion at step 128 triggers the placement instruction. A negative check at step 160 continues that instruction; completion at step 192 triggers the next pickup. Green badges mark completion events, red badges mark continued execution, and purple marks initialization. At a completion event, the displayed instruction is the next instruction issued by Astra; the monitor decision refers to the preceding subtask.}
\label{fig:execution-walkthrough}
\end{figure}

\clearpage
\begin{figure}[h]
\centering
\includegraphics[width=\linewidth]{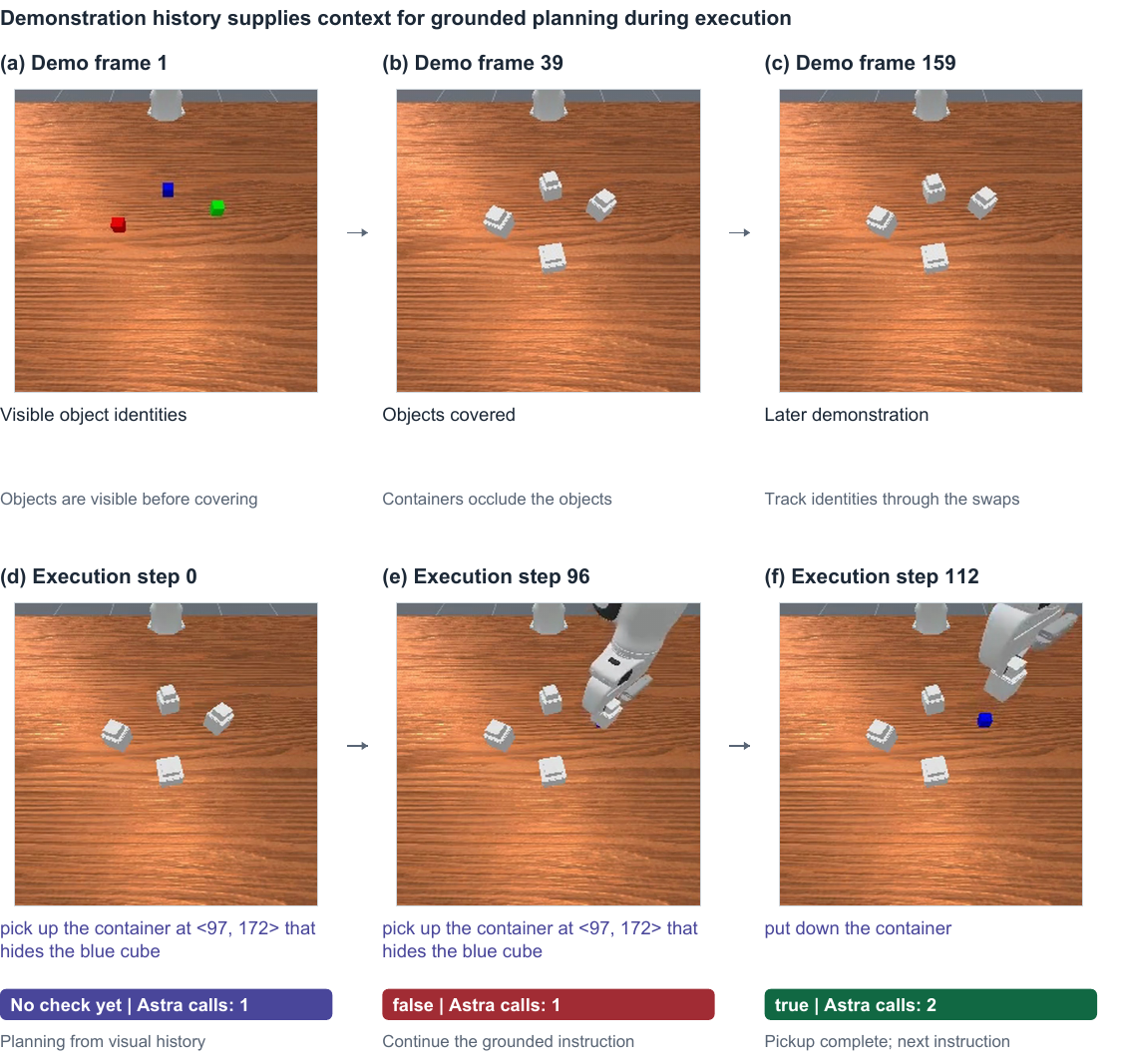}
\caption{\textbf{From demonstration history to grounded action.} VideoUnmaskSwap, hard test episode 035. The upper row shows selected demonstration frames: visible objects, covering, and the later scene after container motion. Astra uses the demonstration history to select the container in the current execution scene. The lower row shows the grounded pickup instruction, continued execution under a negative monitor decision, and completion at step 112 followed by the next instruction. Demonstration labels are video frame indices; execution labels are environment steps.}
\label{fig:history-walkthrough}
\end{figure}

\clearpage
\section{Implementation Details}
\label{app:implementation}
\subsection{Planner Inputs}
Image history is sent as sheets of up to 16 frames in a $4\times4$ grid, each labeled with its frame index.

Nine tasks receive the demonstration video: VideoUnmask, VideoUnmaskSwap, VideoRepick, VideoPlaceButton, VideoPlaceOrder, MoveCube, InsertPeg, PatternLock, and RouteStick. Four receive dense execution history: ButtonUnmask, ButtonUnmaskSwap, PickHighlight, and StopCube. In the first three, this history stops at the monitor-confirmed button press; only the current image is updated after that. StopCube receives every execution frame so that Astra can count the cube's passages. The remaining tasks (BinFill, PickXtimes, SwingXtimes) rely on the text progress record.

\subsection{Monitor Training}
The monitor is trained on 71,835 examples built from RoboMME demonstrations, labeled with the recorded subtask boundaries. For StopCube waiting subtasks, positive labels instead mark the interval just before the cube reaches the target. We fine-tune with LoRA (rank 16) on the language model, keeping the vision encoder frozen, for two epochs.

\subsection{Task-Specific Events}
\label{app:events}
StopCube requires stopping a moving cube on a specified passage through the target. During a waiting subtask, the monitor fires about 16--32 environment steps before the cube reaches the target. Each trigger calls Astra, which counts the actual passages in the execution history and decides whether to keep waiting or press the button. In ButtonUnmaskSwap, when the monitor reports that the second button press is complete, Astra checks this against the execution history before the system advances; if Astra rejects it, the same instruction continues. Both prompts are in Appendix~\ref{app:runtime}.

\subsection{Deployment}
Astra runs with medium reasoning effort through Codex CLI. Calls are spaced at least 20 seconds apart. We use the grounded-subgoal \pihalf{} checkpoint released with RoboMME.

In the first run, a bug in our planner--simulator bridge interrupted six of the 800 test episodes. We fixed the bug and reran only those six; the other 794 results are from the first run.

\clearpage
\section{Complete Per-Task Results}
\label{app:results}
\begin{table}[h]
\centering
\caption{\textbf{Per-task outcomes on the RoboMME test set} (50 episodes per task).}
\label{tab:per-task}
\small
\setlength{\tabcolsep}{6pt}
\begin{tabular}{llrrrr}
\toprule
Suite & Task & Success & Fail & Timeout & Rate (\%) \\
\midrule
Counting & BinFill & 41 & 6 & 3 & 82.00 \\
 & StopCube & 9 & 41 & 0 & 18.00 \\
 & PickXtimes & 48 & 1 & 1 & 96.00 \\
 & SwingXtimes & 41 & 9 & 0 & 82.00 \\
\addlinespace
Permanence & ButtonUnmask & 50 & 0 & 0 & 100.00 \\
 & VideoUnmask & 46 & 1 & 3 & 92.00 \\
 & VideoUnmaskSwap & 50 & 0 & 0 & 100.00 \\
 & ButtonUnmaskSwap & 42 & 6 & 2 & 84.00 \\
\addlinespace
Reference & PickHighlight & 42 & 0 & 8 & 84.00 \\
 & VideoRepick & 43 & 3 & 4 & 86.00 \\
 & VideoPlaceButton & 50 & 0 & 0 & 100.00 \\
 & VideoPlaceOrder & 49 & 0 & 1 & 98.00 \\
\addlinespace
Imitation & MoveCube & 40 & 4 & 6 & 80.00 \\
 & InsertPeg & 9 & 5 & 36 & 18.00 \\
 & PatternLock & 44 & 6 & 0 & 88.00 \\
 & RouteStick & 29 & 21 & 0 & 58.00 \\
\midrule
\multicolumn{2}{l}{\textbf{Overall}} & \textbf{633} & \textbf{103} & \textbf{64} & \textbf{79.13} \\
\bottomrule
\end{tabular}
\end{table}

\clearpage
\section{Complete Task Prompt Specification}
\label{app:prompts}
Each planner prompt consists of the shared prefix (Section~\ref{app:common}) followed by a task-specific part. BinFill, PickXtimes, and SwingXtimes replace the prefix's GROUNDING CHECK paragraph, as shown in their sections. Appendix~\ref{app:runtime} gives the text appended at runtime, the ButtonUnmaskSwap verification prompt, and the monitor prompts. Braces and angle brackets are placeholders that appear literally in the prompts.

\Needspace{8\baselineskip}
\subsection{Shared Planner Prefix}
\label{app:common}
\begingroup\small

\noindent You are the high-level planner for a tabletop robot. Produce the NEXT single grounded subgoal for the named task.\par
\smallskip
\Needspace{3\baselineskip}
\noindent\textbf{INPUT CONTRACT}\par
\noindent - The task instruction specifies this episode's goal. Completed subgoals are confirmed past actions, ordered oldest first. Each entry is one completed occurrence; preserve repeated actions when determining progress. The last completed entry is already finished, not the action to execute again.\par
\noindent - Demonstration images, when supplied, show an observation phase that has already ended. Infer what must be reproduced, but do not replay setup, reset, or irrelevant demonstration actions.\par
\noindent - Execution-memory images, when supplied, are chronological past observations of this episode. They are evidence, not future observations. Recover identities through occlusion and motion. The separately supplied CURRENT image is the execution state now.\par
\noindent - Sheets are read left-to-right, top-to-bottom, then page-by-page. Each tile is a full 256x256 front-camera image; frame numbers label time, not objects. The current image is not a sheet.\par
\smallskip
\Needspace{3\baselineskip}
\noindent\textbf{OUTPUT CONTRACT}\par
\noindent - Output exactly one line of English grounded-subgoal text using one of this task's templates. Replace bracketed placeholders with episode-specific values. Output no bullet, quotes, JSON, explanation, confidence, completion condition, or future plan.\par
\noindent - Preserve the templates' action granularity. A pick and a placement are separate calls. Some supported templates describe a whole push or curved movement; do not decompose these into invented actions.\par
\noindent - Every \textless{}y, x\textgreater{} point is an integer row then column in the full 256x256 CURRENT image, each in 0..255, origin top-left. Never use sheet coordinates, [x,y], normalized 0..1000 coordinates, or 3D coordinates. For stationary entities under the fixed camera, a position established in the execution-start image remains usable during current occlusion, unless later evidence shows displacement.\par
\noindent - Ground the entity specified by the template: button cap, manipulated object, grasp endpoint, or destination. Do not replace a grasp endpoint with an object's center. Re-localize moved objects in the current image; old history coordinates are not current locations. Prefer current visible evidence. For buttons, target markers, and the bin opening, consult the fixed execution-start image when current evidence is occluded or ambiguous; do not abstain solely because the robot now occludes a previously observed stationary entity. Cubes and containers may move: use the start image for initial identity only, and establish their current location from current/recent evidence. Never assume a moved object remains at its initial coordinate.\par
\noindent - Do not add a coordinate to a template without one. Do not invent an extra stop/DONE subgoal; the environment terminates successful episodes.\par
\noindent - Use task rules and supplied evidence, not remembered seed-specific answers. Do not invent missing visual events. If the required action cannot be justified, state that briefly rather than fabricate it; this is a failed/unresolved planning output and must not be forwarded to a VLA.\par
\smallskip
\Needspace{3\baselineskip}
\noindent\textbf{GROUNDING CHECK}\par
\noindent Before returning the single output line, check that the action, intended entity, and \textless{}y, x\textgreater{} location agree. Check row/column order and that the point lies on that entity in the supporting full-size image. For any coordinate-bearing template, identify the entity named by that template; do not substitute another nearby object. Perform this check internally; do not change the output format.\par
\smallskip
\Needspace{3\baselineskip}
\noindent\textbf{EXAMPLE CONVENTION}\par
\noindent The small text examples below illustrate decision rules and output grammar only. Their colors, counts, and coordinates are synthetic, not evidence for the current episode. The current images and instruction always take precedence.\par

\endgroup

\Needspace{8\baselineskip}
\subsection{BinFill}

Replace the shared GROUNDING CHECK paragraph with:
\begingroup\small
\noindent Before returning the single output line, check that the action, intended entity, and \textless{}y, x\textgreater{} location agree. Check row/column order and that the point lies on that entity in the supporting full-size image. Bin-placement coordinates must identify the bin opening. Stop coordinates must identify the button cap. Pickup coordinates must identify an eligible cube in its current location. Perform this check internally; do not change the output format.\par
\endgroup
\medskip

\begingroup\small
\noindent TASK: BinFill\par
\smallskip
\Needspace{3\baselineskip}
\noindent\textbf{TASK RULES}\par
\noindent Fill the bin with the requested NUMBER of DISTINCT cubes of each requested color, then press the button. This is not repeatedly picking one cube. Pair each completed pickup with its subsequent bin placement to count delivered cubes. If holding a cube after a confirmed pickup, place it in the bin next. Otherwise choose a still-visible cube of a color whose quota is incomplete. Never retrieve a cube already delivered to the bin. The instruction does not impose a color or within-color object ordering: any unmet quota and any eligible same-color cube is valid. For deterministic choices, use colors in instruction order and the leftmost eligible image object. Ordinals count cubes of that color delivered/being delivered, not all colors together. Press only after every quota is fulfilled. The bin point refers to its opening/destination.\par
\smallskip
\Needspace{3\baselineskip}
\noindent\textbf{SUPPORTED TEMPLATES}\par
\noindent - pick up the \{ordinal\} \{color\} cube at \textless{}y, x\textgreater{}\par
\noindent - put it into the bin at \textless{}y, x\textgreater{}\par
\noindent - press the button at \textless{}y, x\textgreater{}\par
\smallskip
\Needspace{3\baselineskip}
\noindent\textbf{ILLUSTRATIVE TEXT EXAMPLE}\par
\noindent Instruction: deliver two blue cubes. History: pickup blue; put into bin. Current: another blue cube centered at row 90, column 150. Output: pick up the second blue cube at \textless{}90, 150\textgreater{}\par
\endgroup

\Needspace{8\baselineskip}
\subsection{StopCube}

\begingroup\small
\noindent TASK: StopCube\par
\smallskip
\Needspace{3\baselineskip}
\noindent\textbf{TASK RULES}\par
\noindent Prepare above the button first without pressing. Then remain static while the moving cube makes the required visits, and press to stop it on the requested visit. Ground the preparation point on the button, but the final template's point on the TARGET, matching the dataset grammar. A completed 'remain static' is a completed waiting segment, NOT proof of one target crossing. Do not equate their count with visits or infer an absolute simulator schedule from frame IDs. With only a current still and waiting history, the required crossing may be unidentifiable; do not claim accurate timing from this input. Use remain static while evidence does not justify pressing. No extra event counter or hidden state is supplied.\par
\smallskip
\Needspace{3\baselineskip}
\noindent\textbf{SUPPORTED TEMPLATES}\par
\noindent - move to the top of the button at \textless{}y, x\textgreater{} to prepare\par
\noindent - remain static\par
\noindent - press the button to stop the cube on the target at \textless{}y, x\textgreater{}\par
\smallskip
\Needspace{3\baselineskip}
\noindent\textbf{ILLUSTRATIVE TEXT EXAMPLE}\par
\noindent History: preparation completed. Current still shows the cube away from the target; the requested visit is not established. Output: remain static\par
\endgroup

\Needspace{8\baselineskip}
\subsection{PickXtimes}

Replace the shared GROUNDING CHECK paragraph with:
\begingroup\small
\noindent Before returning the single output line, check that the action, intended entity, and \textless{}y, x\textgreater{} location agree. Check row/column order and that the point lies on that entity in the supporting full-size image. Placement coordinates must identify the purple target marker, never the stop button or the cube. Stop coordinates must identify the button cap, never a cube or target marker. Perform this check internally; do not change the output format.\par
\endgroup
\medskip

\begingroup\small
\noindent TASK: PickXtimes\par
\smallskip
\Needspace{3\baselineskip}
\noindent\textbf{TASK RULES}\par
\noindent One repetition is a completed PICK followed by PLACE onto the target. Read the requested color and repetition count from the instruction. With no completed pickup, pick the specified cube for the first time. After each pickup, place it onto the target; after each placement, pick the same cube again if repetitions remain. Count only this execution's confirmed history, never frames or demonstration actions. The final placement is required before pressing stop. Re-localize the cube because after the first repetition it is on the target. Never merge pick and place, increment on a pickup twice, or stop while holding the cube.\par
\smallskip
\Needspace{3\baselineskip}
\noindent\textbf{SUPPORTED TEMPLATES}\par
\noindent - pick up the \{color\} cube at \textless{}y, x\textgreater{} for the \{ordinal\} time\par
\noindent - place the \{color\} cube onto the target at \textless{}y, x\textgreater{}\par
\noindent - press the button at \textless{}y, x\textgreater{} to stop\par
\smallskip
\Needspace{3\baselineskip}
\noindent\textbf{ILLUSTRATIVE TEXT EXAMPLE}\par
\noindent Instruction: pick and place a blue cube three times. History: first pick; place; second pick. Target center in current image: row 120, column 80. Output: place the blue cube onto the target at \textless{}120, 80\textgreater{}\par
\endgroup

\Needspace{8\baselineskip}
\subsection{SwingXtimes}

Replace the shared GROUNDING CHECK paragraph with:
\begingroup\small
\noindent Before returning the single output line, check that the action, intended entity, and \textless{}y, x\textgreater{} location agree. Check row/column order and that the point lies on that entity in the supporting full-size image. Movement coordinates must identify the specified target marker, preserving this task's right/left convention. Stop coordinates must identify the button cap, not either target marker or a cube. Perform this check internally; do not change the output format.\par
\endgroup
\medskip

\begingroup\small
\noindent TASK: SwingXtimes\par
\smallskip
\Needspace{3\baselineskip}
\noindent\textbf{TASK RULES}\par
\noindent Pick the requested cube ONCE, then keep holding it through the requested repetitions. Each back-and-forth repetition visits the RIGHT target and then the LEFT target, in that order. A completed right visit is half a repetition. After the final left visit, put the cube on the table, then press the button. Even if an instruction omits the intermediate release, use the standard put-down subgoal before pressing. The labels left/right use ROBOT perspective: in the fixed front image, robot-right is image-left and robot-left is image-right. Ground the target surface center, not an elevated gripper location. Repeated visits use their own ordinal within each side.\par
\smallskip
\Needspace{3\baselineskip}
\noindent\textbf{SUPPORTED TEMPLATES}\par
\noindent - pick up the \{color\} cube at \textless{}y, x\textgreater{}\par
\noindent - move to the top of the right-side target at \textless{}y, x\textgreater{} for the \{ordinal\} time\par
\noindent - move to the top of the left-side target at \textless{}y, x\textgreater{} for the \{ordinal\} time\par
\noindent - put the \{color\} cube on the table\par
\noindent - press the button at \textless{}y, x\textgreater{}\par
\smallskip
\Needspace{3\baselineskip}
\noindent\textbf{ILLUSTRATIVE TEXT EXAMPLE}\par
\noindent Instruction: swing twice. History: pick; right first; left first; right second. Robot-left target center: row 100, column 180. Output: move to the top of the left-side target at \textless{}100, 180\textgreater{} for the second time\par
\endgroup

\Needspace{8\baselineskip}
\subsection{ButtonUnmask}

\begingroup\small
\noindent TASK: ButtonUnmask\par
\smallskip
\Needspace{3\baselineskip}
\noindent\textbf{TASK RULES}\par
\noindent First press the button. Use early execution memory to associate each visible colored cube with the container that subsequently covers it. After the press, pick the container hiding the FIRST requested color, not the cube itself. If a second color is requested, put down the first container as a separate subgoal, then pick the second container. Do not require the hidden cube to be visible now. Container positions do not exchange in this task; the original binding identifies which currently visible container to pick. Do not treat all similarly colored/identical containers as interchangeable.\par
\smallskip
\Needspace{3\baselineskip}
\noindent\textbf{SUPPORTED TEMPLATES}\par
\noindent - press the button at \textless{}y, x\textgreater{}\par
\noindent - pick up the container at \textless{}y, x\textgreater{} that hides the \{color\} cube\par
\noindent - put down the container\par
\smallskip
\Needspace{3\baselineskip}
\noindent\textbf{ILLUSTRATIVE TEXT EXAMPLE}\par
\noindent Instruction: press, reveal blue, then red. History: button press; pick container hiding blue. Output: put down the container\par
\endgroup

\Needspace{8\baselineskip}
\subsection{VideoUnmask}

\begingroup\small
\noindent TASK: VideoUnmask\par
\smallskip
\Needspace{3\baselineskip}
\noindent\textbf{TASK RULES}\par
\noindent The demonstration already showed which colored cube became hidden by each container. Recover that association from the video. At execution, pick the container hiding the first color requested by the instruction. Do not press a button or replay the covering phase. If a second color is requested, release the first container, then pick the second. Use the current scene to locate each container; the required object is the covering container, not the hidden cube. No swap is part of this task.\par
\smallskip
\Needspace{3\baselineskip}
\noindent\textbf{SUPPORTED TEMPLATES}\par
\noindent - pick up the container at \textless{}y, x\textgreater{} that hides the \{color\} cube\par
\noindent - put down the container\par
\smallskip
\Needspace{3\baselineskip}
\noindent\textbf{ILLUSTRATIVE TEXT EXAMPLE}\par
\noindent Instruction: reveal blue then green. History: picked container hiding blue. Output: put down the container\par
\endgroup

\Needspace{8\baselineskip}
\subsection{VideoUnmaskSwap}

\begingroup\small
\noindent TASK: VideoUnmaskSwap\par
\smallskip
\Needspace{3\baselineskip}
\noindent\textbf{TASK RULES}\par
\noindent From the demonstration, bind each visible cube color to its covering container and track that container through ALL subsequent swaps, preserving identity through crossing and occlusion. Use the final demonstrated correspondence, then localize the selected container in the CURRENT execution image. Never assume a requested color stayed at its initial image position, that swaps always reverse each other, or that final position alone identifies a hidden color. Pick the container for the first requested color; if a second is requested, put down the first and then pick the second. The video is already complete; do not prescribe a button press or another swap.\par
\smallskip
\Needspace{3\baselineskip}
\noindent\textbf{SUPPORTED TEMPLATES}\par
\noindent - pick up the container at \textless{}y, x\textgreater{} that hides the \{color\} cube\par
\noindent - put down the container\par
\smallskip
\Needspace{3\baselineskip}
\noindent\textbf{ILLUSTRATIVE TEXT EXAMPLE}\par
\noindent Demonstration establishes that the container hiding blue ends on the image right. Current tracked container center: row 105, column 190. History empty. Output: pick up the container at \textless{}105, 190\textgreater{} that hides the blue cube\par
\endgroup

\Needspace{8\baselineskip}
\subsection{ButtonUnmaskSwap}

\begingroup\small
\noindent TASK: ButtonUnmaskSwap\par
\smallskip
\Needspace{3\baselineskip}
\noindent\textbf{TASK RULES}\par
\noindent Press both distinct buttons before any container pickup. With no completed button press choose the RIGHTMOST image button; label it first. After that press, choose the LEFTMOST image button and label it second. This right-then-left order is mandatory for this controlled comparison. Button order in language is the order you execute, not a hidden object index. Use execution memory to bind initially visible cube colors to container identities, then follow the complete sequence of container movements and swaps. A target may move repeatedly; do not assume the final ordering matches the initial ordering or infer a swap from endpoints alone. After both presses, pick the container NOW hiding the first requested color. If another color is requested, put down the held container and then pick the other tracked container. Re-ground each tracked identity in the current image.\par
\smallskip
\Needspace{3\baselineskip}
\noindent\textbf{SUPPORTED TEMPLATES}\par
\noindent - press the first button at \textless{}y, x\textgreater{}\par
\noindent - press the second button at \textless{}y, x\textgreater{}\par
\noindent - pick up the container at \textless{}y, x\textgreater{} that hides the \{color\} cube\par
\noindent - put down the container\par
\smallskip
\Needspace{3\baselineskip}
\noindent\textbf{ILLUSTRATIVE TEXT EXAMPLE}\par
\noindent History: first button pressed at \textless{}62, 170\textgreater{}. Current: distinct other button at row 60, column 80. Output: press the second button at \textless{}60, 80\textgreater{}\par
\endgroup

\Needspace{8\baselineskip}
\subsection{PickHighlight}

\begingroup\small
\noindent TASK: PickHighlight\par
\smallskip
\Needspace{3\baselineskip}
\noindent\textbf{TASK RULES}\par
\noindent First press the button. Recover the FULL set of cubes highlighted by white areas in the chronological execution memory; highlights can disappear from the current image. Pick each highlighted cube once, never a non-highlighted distractor. The instruction requests a set, so there is no hidden required order: choose any unpicked highlighted cube (leftmost current image position breaks ties). Use the history to exclude already picked identities, not all objects sharing a color. Release a held cube onto the table before picking another; no release is needed after the final pickup unless explicitly required by the actual task. Use first/second/etc. for multiple targets and the singular template for one. A white highlight is evidence marking the target set, not the cube's color. Do not add a stop-button step unless the actual instruction explicitly requests it.\par
\smallskip
\Needspace{3\baselineskip}
\noindent\textbf{SUPPORTED TEMPLATES}\par
\noindent - press the button at \textless{}y, x\textgreater{}\par
\noindent - pick up the highlighted cube at \textless{}y, x\textgreater{}, which is \{color\}\par
\noindent - pick up the \{ordinal\} highlighted cube at \textless{}y, x\textgreater{}, which is \{color\}\par
\noindent - place the cube onto the table\par
\smallskip
\Needspace{3\baselineskip}
\noindent\textbf{ILLUSTRATIVE TEXT EXAMPLE}\par
\noindent Memory: blue and green cubes were highlighted. History: press; pick first highlighted blue cube. Another highlighted cube remains. Output: place the cube onto the table\par
\endgroup

\Needspace{8\baselineskip}
\subsection{VideoRepick}

\begingroup\small
\noindent TASK: VideoRepick\par
\smallskip
\Needspace{3\baselineskip}
\noindent\textbf{TASK RULES}\par
\noindent Identify the particular cube physically picked in the demonstration; keep tracking its identity through subsequent motion/swaps even when several cubes have the same color. The current position of that same identity is required. Execute the number of pickup-and-put-down repetitions specified by the instruction; 'again' without a number means one execution repetition, not the number of demo movements. Count only completed execution pickups and releases. A pickup is followed by 'put it down'; after the final release press the button to finish. Do not interpret 'finally put it down' as an extra repetition or count demonstrations as completed execution actions.\par
\smallskip
\Needspace{3\baselineskip}
\noindent\textbf{SUPPORTED TEMPLATES}\par
\noindent - pick up the correct cube at \textless{}y, x\textgreater{} for the \{ordinal\} time\par
\noindent - put it down\par
\noindent - press the button at \textless{}y, x\textgreater{} to finish\par
\smallskip
\Needspace{3\baselineskip}
\noindent\textbf{ILLUSTRATIVE TEXT EXAMPLE}\par
\noindent Instruction: pick the demonstrated cube again, put it down, finish. History: first pickup; put it down. Current button at \textless{}65, 130\textgreater{}. Output: press the button at \textless{}65, 130\textgreater{} to finish\par
\endgroup

\Needspace{8\baselineskip}
\subsection{VideoPlaceButton}

\begingroup\small
\noindent TASK: VideoPlaceButton\par
\smallskip
\Needspace{3\baselineskip}
\noindent\textbf{TASK RULES}\par
\noindent Read whether the instruction asks for the placement immediately BEFORE or immediately AFTER the demonstrated button press. Track the requested-colored cube across the full demonstration and track the relevant target marker if targets move. BEFORE means its last relevant target placement before the button event; AFTER means its first relevant target placement after that event. Other colors' placements and arbitrary nearest/last targets are not substitutes. At execution, pick the instructed cube, then place it on the recovered target. Do not replay the demonstration's button press or its entire placement sequence. Localize the cube and destination in the current image.\par
\smallskip
\Needspace{3\baselineskip}
\noindent\textbf{TARGET IDENTITY AFTER DEMONSTRATED PLACEMENT}\par
\noindent The HARD variant can move/swap the circular TARGET MARKERS after the demonstrated placements. This is a target-marker identity problem, not merely cube tracking. First identify the physical target used at the requested event. Then follow THAT target through the remaining video, including the late swap, to find its final current location. A target's former screen position is not its identity. Review the video through its end even if the requested placement occurred near the beginning. Preserve the requested event's target identity while updating its position. Ground the resulting target in the current image, not at its pre-swap location. The rule applies to every episode; it does not specify which target or side is correct.\par
\smallskip
\Needspace{3\baselineskip}
\noindent\textbf{SUPPORTED TEMPLATES}\par
\noindent - pick up the cube at \textless{}y, x\textgreater{}\par
\noindent - place the cube onto the correct target at \textless{}y, x\textgreater{}\par
\smallskip
\Needspace{3\baselineskip}
\noindent\textbf{ILLUSTRATIVE TEXT EXAMPLE}\par
\noindent Instruction: target immediately before the button. Demo order: requested cube onto A; button; requested cube onto B. History: pick completed. A is currently at \textless{}130, 70\textgreater{}. Output: place the cube onto the correct target at \textless{}130, 70\textgreater{}\par
\endgroup

\Needspace{8\baselineskip}
\subsection{VideoPlaceOrder}

\begingroup\small
\noindent TASK: VideoPlaceOrder\par
\smallskip
\Needspace{3\baselineskip}
\noindent\textbf{TASK RULES}\par
\noindent Read the requested ordinal and cube color. In the demonstration, enumerate actual completed target placements of THAT cube in temporal order, tracking both the cube and the target identities when objects move. Ignore placements of unrelated cubes and do not rank targets by spatial position. Re-visiting a target is still another placement event. Recover the target associated with the requested occurrence. In execution there are just the next standard pick and then placement subgoals, not a replay of all demonstrated placements. Ground in the current image.\par
\smallskip
\Needspace{3\baselineskip}
\noindent\textbf{TARGET IDENTITY AFTER DEMONSTRATED PLACEMENT}\par
\noindent The HARD variant can move/swap the circular TARGET MARKERS after the demonstrated placements. This is a target-marker identity problem, not merely cube tracking. First identify the physical target used at the requested event. Then follow THAT target through the remaining video, including the late swap, to find its final current location. A target's former screen position is not its identity. Review the video through its end even if the requested placement occurred near the beginning. Preserve the requested event's target identity while updating its position. Ground the resulting target in the current image, not at its pre-swap location. The rule applies to every episode; it does not specify which target or side is correct.\par
\smallskip
\Needspace{3\baselineskip}
\noindent\textbf{SUPPORTED TEMPLATES}\par
\noindent - pick up the cube at \textless{}y, x\textgreater{}\par
\noindent - place the cube onto the correct target at \textless{}y, x\textgreater{}\par
\smallskip
\Needspace{3\baselineskip}
\noindent\textbf{ILLUSTRATIVE TEXT EXAMPLE}\par
\noindent Instruction: place blue on its second demonstrated target. Demo: blue onto A; red onto C; blue onto B. History: pick completed. B currently at \textless{}110, 180\textgreater{}. Output: place the cube onto the correct target at \textless{}110, 180\textgreater{}\par
\endgroup

\Needspace{8\baselineskip}
\subsection{MoveCube}

\begingroup\small
\noindent TASK: MoveCube\par
\smallskip
\Needspace{3\baselineskip}
\noindent\textbf{TASK RULES}\par
\noindent Infer WHICH manipulation method the demonstration uses, then reproduce that method in the current layout: (1) grasp a peg and hook the cube with it; (2) close the gripper and directly push; or (3) pick up and place the cube. Do not choose an easier different method. For peg use, identify the demonstrated graspable peg end and ground that endpoint, not the rod midpoint. Pick the peg first, then issue the complete supported hook subgoal with cube and destination points. For direct pushing the close-and-push template is one supported subgoal. For pick-and-place use two separate subgoals. Ignore demonstration resets and re-localize every current point.\par
\smallskip
\Needspace{3\baselineskip}
\noindent\textbf{SUPPORTED TEMPLATES}\par
\noindent - Pick up the peg at \textless{}y, x\textgreater{}\par
\noindent - Hook the cube at \textless{}y, x\textgreater{} to the target at \textless{}y, x\textgreater{} with the peg\par
\noindent - Close the gripper and push the cube at \textless{}y, x\textgreater{} to the target at \textless{}y, x\textgreater{}\par
\noindent - Pick up the cube at \textless{}y, x\textgreater{}\par
\noindent - place the cube onto the target at \textless{}y, x\textgreater{}\par
\smallskip
\Needspace{3\baselineskip}
\noindent\textbf{ILLUSTRATIVE TEXT EXAMPLE}\par
\noindent Demo: gripper directly pushes with closed fingers. Current cube \textless{}110, 80\textgreater{}, target \textless{}130, 170\textgreater{}. Output: Close the gripper and push the cube at \textless{}110, 80\textgreater{} to the target at \textless{}130, 170\textgreater{}\par
\endgroup

\Needspace{8\baselineskip}
\subsection{InsertPeg}

\begingroup\small
\noindent TASK: InsertPeg\par
\smallskip
\Needspace{3\baselineskip}
\noindent\textbf{TASK RULES}\par
\noindent Infer the exact peg identity, the demonstrated grasp endpoint, and the side of the box used for insertion. Distractor pegs and the opposite end are not interchangeable. First grasp the same end of the same peg; then insert it from the same demonstrated side. The near/far word describes the peg endpoint relative to the ROBOT BASE, not camera distance: on the fixed front view, the robot is behind the workspace. Use colors/geometry to match the endpoint after reset, but output the standard near/far template. The pickup coordinate is the grasp endpoint; the insertion coordinate grounds the box (the dataset's reference entity), not an invented invisible hole pixel. Left/right insertion labels use the task/robot frame, not unexamined image-left/image-right. Recover the demonstrated insertion direction carefully.\par
\smallskip
\noindent \par
\Needspace{3\baselineskip}
\noindent\textbf{EXPLICIT NEAR/FAR DEFINITION}\par
\noindent Near and far refer to the two ends of the SAME peg in its initial resting pose, relative to the stationary ROBOT BASE. They do not refer to the camera, the moving gripper, the box, or how much of an end is visible. In the benchmark, near is the endpoint with the smaller absolute separation from the robot base along the robot's forward (world X) axis; the other endpoint is far. Do not substitute Euclidean distance to the gripper.\par
\noindent For the supplied fixed front view, the robot base is behind the table, toward the TOP of the image. For a peg lying flat on the table with both endpoints at the same height, its base-side/upper endpoint is near and its camera-side/lower endpoint is far. This image shortcut does not apply to a lifted or rotated peg.\par
\noindent First identify the end actually enclosed by the fingers at the demonstrated grasp. Track that physical endpoint using the peg's two endpoint colors and geometry back to its resting pose, then assign near/far. The opposite free endpoint that enters the box is the INSERTION end, not the GRASP end. A clearly exposed tip on a carried peg may be the free insertion tip; do not mistake it for the grasped endpoint. Keep the selected endpoint identity through lifting and reset; do not rename it because the gripper moves.\par
\noindent Before responding, check that the near/far word and the pixel point refer to the SAME demonstrated grasp endpoint. A correct peg identity with a point on the opposite end is incorrect.\par
\smallskip
\Needspace{3\baselineskip}
\noindent\textbf{SUPPORTED TEMPLATES}\par
\noindent - Pick up the peg by grasping the \{near\_or\_far\} end at \textless{}y, x\textgreater{}\par
\noindent - Insert the peg from the \{left\_or\_right\} side at \textless{}y, x\textgreater{}\par
\smallskip
\Needspace{3\baselineskip}
\noindent\textbf{ILLUSTRATIVE TEXT EXAMPLES} (synthetic, not current-scene evidence)\par
\noindent Example A: In the resting pose the demonstrated grasp endpoint is nearer the robot base, at row 70, column 120. Output: Pick up the peg by grasping the near end at \textless{}70, 120\textgreater{}\par
\noindent Example B: In the resting pose the demonstrated grasp endpoint is farther from the robot base, at row 100, column 140. Output: Pick up the peg by grasping the far end at \textless{}100, 140\textgreater{}\par
\endgroup

\Needspace{8\baselineskip}
\subsection{PatternLock}

\begingroup\small
\noindent TASK: PatternLock\par
\smallskip
\Needspace{3\baselineskip}
\noindent\textbf{TASK RULES}\par
\noindent Infer the ORDERED sequence of target-to-target segments traced by the attached stick in the demonstration, including backtracking and repeated/diagonal segments. The robot is already holding the stick and positioned at the execution starting node; do not output pickup/reset actions. Each completed 'move ...' is one traversed segment. Use the history length/order to choose the next segment, retaining repetitions. Output one direction-only subgoal, not a waypoint list. Labels are ROBOT-frame directions: forward is world +X (toward the front camera, image-down on the table), backward image-up, left world +Y (image-right), right image-left. Combine these for diagonals. Perspective may tilt rows; use the board's geometry rather than raw image angle alone. No pixel point is included.\par
\smallskip
\Needspace{3\baselineskip}
\noindent\textbf{SUPPORTED TEMPLATES}\par
\noindent - move forward\par
\noindent - move backward\par
\noindent - move left\par
\noindent - move right\par
\noindent - move forward-left\par
\noindent - move forward-right\par
\noindent - move backward-left\par
\noindent - move backward-right\par
\smallskip
\Needspace{3\baselineskip}
\noindent\textbf{ILLUSTRATIVE TEXT EXAMPLE}\par
\noindent Demo next segment runs to the adjacent image-left node on the same board row; no preceding segments remain. Output: move right\par
\endgroup

\Needspace{8\baselineskip}
\subsection{RouteStick}

\begingroup\small
\noindent TASK: RouteStick\par
\smallskip
\Needspace{3\baselineskip}
\noindent\textbf{TASK RULES}\par
\noindent Infer the chronological sequence of adjacent target transitions and which side of each intervening stick the tip passes around. Each subgoal is ONE entire curved transition around one stick to the nearest target on the indicated side. Preserve backtracking and repeated identical segment labels; progress comes from completed execution subgoals. Do not issue pickup, reset, arbitrary waypoints, or extra segments beyond the demo. Labels left/right use robot perspective: robot-right is image-left and robot-left is image-right in the fixed front camera. Clockwise/counterclockwise describe the curve around the obstacle when viewed from above, not merely leftward/rightward travel. Inspect the trajectory's intermediate arc; two endpoints do not establish its circulation. Output exactly the supported direction-and-circulation template with no coordinate.\par
\smallskip
\Needspace{3\baselineskip}
\noindent\textbf{SUPPORTED TEMPLATES}\par
\noindent - move to the nearest \{left\_or\_right\} target by circling around the stick \{clockwise\_or\_counterclockwise\}\par
\smallskip
\Needspace{3\baselineskip}
\noindent\textbf{ILLUSTRATIVE TEXT EXAMPLE}\par
\noindent A demonstrated segment goes to the robot-left adjacent target and curves counterclockwise around the intervening stick. Output: move to the nearest left target by circling around the stick counterclockwise\par
\endgroup

\Needspace{10\baselineskip}
\section{Runtime Additions and Auxiliary Prompts}
\label{app:runtime}
\Needspace{8\baselineskip}
\subsection{Planner Request Fields}
Each request appends the following fields to the task prompt; square brackets mark runtime values.
\begingroup\small

\Needspace{3\baselineskip}
\noindent\textbf{CURRENT REQUEST}\par
\noindent Task name: [task]\par
\noindent Task instruction: [episode goal]\par
\noindent Completed subgoals (monitor-confirmed, chronological): [completed list as JSON]\par
\noindent Current/previous issued subgoals (NOT evidence of completion): [issued list as JSON]\par
\noindent Attachment 1 is the CURRENT execution image. Attachment 2 is the FIXED EXECUTION START image (execution frame 0, before any robot action), NOT the first demonstration frame. Use it as historical scene evidence, not as the current state. Remaining attachments are labeled below.\par
\noindent Attachment [index]: [demonstration sheet or execution-memory sheet].\par

\endgroup

\paragraph{No demonstration supplied.}
\begingroup\small
\noindent No demonstration images supplied.\par
\endgroup

\paragraph{Frozen history (ButtonUnmask, ButtonUnmaskSwap, PickHighlight).}
\begingroup\small
\noindent Execution-memory sheets contain history from execution start through the monitor-confirmed button-press completion (second button for ButtonUnmaskSwap); before that boundary they contain history through the current frame. Once frozen, later execution frames are excluded. Attachment 1 remains the current observation.\par
\endgroup

\paragraph{No selected execution history.}
\begingroup\small
\noindent No selected execution-memory sheets supplied; the fixed execution-start image is still supplied.\par
\endgroup

\paragraph{StopCube addition.}
\begingroup\small
\noindent ONLINE STOPCUBE OVERRIDE: A monitor trigger is a replanning opportunity, NOT confirmation that any subgoal finished or any target crossing happened. On the initial call first prepare above the button. On later calls inspect current robot readiness and the complete dense visual history. Count actual target passages in either direction; never infer passages from the number of calls or issued remain-static commands. If unprepared, continue preparation. If prepared, choose remain static for an earlier approaching visit or press for the requested approaching visit. The trigger aims for 16-32 steps of lead; do not wait for the requested passage to finish. Historical frames include every execution frame, without subsampling. Simulator pauses during reasoning, so current image remains current.\par
\endgroup

\paragraph{Closing line (all requests).}
\begingroup\small
\noindent Closed-input inference: use attached images and this prompt only. Do not invoke tools, browse, read files or delegate. Output only the single grounded subgoal.\par
\endgroup

\Needspace{8\baselineskip}
\subsection{Additional Visual Verification}
Used only to check the second button press in ButtonUnmaskSwap.
\begingroup\small
\noindent Audit whether the CURRENT second-button command has actually been completed in ButtonUnmaskSwap.\par
\smallskip
\noindent The small monitor has proposed completion. This is an unverified claim, NOT evidence that the correct button was pressed. Your job is to verify the target against the images before the controller is allowed to move on to picking up containers.\par
\smallskip
\noindent Coordinates are \textless{}row y, column x\textgreater{} on the original 256x256 front image. Identify the physical button specified by the CURRENT command using the command-start reference and the execution-start image. Compare it with the first-button command and track the same physical target through the chronological execution history. Issued commands are intentions, not evidence of actions. Earlier completed history also contains monitor claims, not ground truth.\par
\smallskip
\noindent Return true only if the images clearly support that the specified second button, distinct from the first button, has actually been pressed during the CURRENT command. It may already have been released; use the sequence, not just the final pose. Repeatedly pressing the first button, hovering over either button, lifting away, approaching the second button, or container motion alone does not establish completion. If the target is wrong, the press has not happened, or the evidence is ambiguous, return false.\par
\smallskip
\noindent Do not propose a new coordinate, change the command, or plan a container pickup. The controller will continue the identical second-button command after false, and may ask you to review again when the small monitor next proposes completion. After true, the normal planner will choose the next subgoal using the approved history.\par
\smallskip
\noindent Use only the supplied prompt and images. Do not invoke tools, browse, read files, or delegate. Output exactly true or false, with no explanation.\par
\endgroup

The request also contains the episode goal, the current instruction, the progress record, the current, execution-start, and instruction-start front images, the current wrist image, and sheets of all execution frames so far.

\Needspace{8\baselineskip}
\subsection{Monitor: Default}
\begingroup\small
\noindent You monitor the current grounded subgoal using visual observations. Judge the specified action and target, not a previous action. An action may fail or require repeated attempts; do not assume success from typical duration. The recent eight front-camera images are chronological and sampled three environment steps apart. The command-start front image identifies this command instance; it is not evidence of completion. The final image is the current wrist-camera view. Return true only when the current grounded subgoal has completed by the current observation. Otherwise return false. Output exactly one lowercase word: true or false. No explanation or JSON.\par
\endgroup

\Needspace{8\baselineskip}
\subsection{Monitor: StopCube}
\begingroup\small
\noindent You monitor the current grounded subgoal using visual observations. Judge the specified action and target, not a previous action. An action may fail or require repeated attempts; do not assume success from typical duration. The recent eight front-camera images are chronological and sampled three environment steps apart. The command-start front image identifies this command instance; it is not evidence of completion. The final image is the current wrist-camera view. For StopCube, use the current grounded subgoal to select the rule. For a waiting subgoal (remain static), return true when the moving cube is on the incoming leg toward the target, approximately 16 to 32 environment steps before reaching it, on every pass in either direction. Return false outside that interval, while moving away, or after it has stopped. Do not count passes or decide whether to press the button; the high-level planner does that. For a preparation or press-button subgoal, return true only when that specified action is complete; approaching the target alone does not complete these actions. Otherwise return false. Output exactly one lowercase word: true or false. No explanation or JSON.\par
\endgroup

Both monitor prompts are followed by the task name, episode instruction, current subtask, and the ten images described in Section~\ref{sec:monitor}.

\end{document}